\documentclass[conference]{IEEEtran}

\usepackage{makecell}
\usepackage{pifont}   
\usepackage{adjustbox}
\usepackage{pgfplots}
\usepackage{cite}
\usepackage{amsmath,amssymb,amsfonts}
\usepackage{algorithmic}
\usepackage{graphicx}
\usepackage{textcomp}
\usepackage{array}
\usepackage{comment}
\usepackage[nolist]{acronym}
\usepackage[normalem]{ulem}
\usepackage{color,soul}
\usepackage{multirow}
\usepackage{multicol,multirow,colortbl}
\usepackage{enumitem}
\usepackage{algorithm}
\usepackage[dvipsnames]{xcolor}
\usepackage{subcaption}

\UseRawInputEncoding

\begin{acronym} 
\acro{3GPP}{Third Generation Partnership Project}
\acro{5G}{fifth generation}
\acro{API}{application programming interface}
\acro{C-ITS}{cooperative-intelligent transport systems}
\acro{C-V2X}{cellular-vehicle-to-everything}
\acro{CAM}{Cooperative Awareness Message}
\acro{CAV}{connected and autonomous vehicle}
\acro{CCAM}{cooperative connected and automated mobility}
\acro{CPM}{collective perception message}
\acro{DCC}{Decentralized Congestion Control}
\acro{DENM}{decentralized environmental notification message}
\acro{DSRC}{dedicated short range communication}
\acro{ETSI}{European Telecommunications Standards Institute}
\acro{FFT}{fast Fourier transform}
\acro{GNSS}{global navigation satellite system}
\acro{GPS}{global positioning system}
\acro{HAT}{hardware attached-on-top}
\acro{HMI}{human-machine interface}
\acro{ICW}{intersection collision warning}
\acro{IMU}{Inertial Measurement Unit}
\acro{IQ}{in-phase and in-quadrature}
\acro{ISAC}{integrated sensing and communication}
\acro{ITS}{intelligent transport system}
\acro{IVIM}{infrastructure to vehicle information message}
\acro{LDM}{Local Dynamic Map}
\acro{LOS}{line-of-sight}
\acro{MAC}{medium access control}
\acro{MCM}{maneuver coordination message}
\acro{MIMO}{multiple input multiple output}
\acro{NLOS}{non-line-of-sight}
\acro{NMEA}{National Marine Electronics Association}
\acro{OBU}{on-board unit}
\acro{OFDM}{orthogonal frequency-division multiplexing}
\acro{OScar}{Open Stack for Car}
\acro{PHY}{physical}
\acro{PVT}{Position Velocity Time}
\acro{ROS2}{Robot Operating System~2}
\acro{RSU}{road side unit}
\acro{TLM}{Traffic Light Maneuver}
\acro{V2N}{vehicle-to-network}
\acro{V2I}{vehicle-to-infrastructure} 
\acro{V2V}{vehicle-to-vehicle} 
\acro{V2X}{vehicle-to-everything} 
\acro{VAM}{vulnerable road user awareness message}
\acro{VESC}{Vedder Electronic Speed Controller}
\acro{WAVE}{wireless access in vehicular environment}
\end{acronym}

\begin{document}
\bstctlcite{IEEEexample:BSTcontrol}
%
\title{Open-Source Based and ETSI Compliant\\Cooperative, Connected, and Automated Mini-Cars}

\author{\IEEEauthorblockN{
Lorenzo Farina\IEEEauthorrefmark{1}\IEEEauthorrefmark{2}, 
Federico Gavioli\IEEEauthorrefmark{3}, 
Salvatore Iandolo\IEEEauthorrefmark{3}, 
Francesco Moretti\IEEEauthorrefmark{3}, 
Giuseppe Perrone\IEEEauthorrefmark{5},\\ 
Matteo Piccoli\IEEEauthorrefmark{1}, 
Francesco Raviglione\IEEEauthorrefmark{4}, 
Marco Rapelli\IEEEauthorrefmark{5}, 
Antonio Solida\IEEEauthorrefmark{3},\\ 
Paolo Burgio\IEEEauthorrefmark{3}, 
Carlo Augusto Grazia\IEEEauthorrefmark{3}, 
and Alessandro Bazzi\IEEEauthorrefmark{1}\IEEEauthorrefmark{2}
}\\
\IEEEauthorblockA{\IEEEauthorrefmark{1}DEI, Universit\`a di Bologna, 40136 Bologna, Italy}
\IEEEauthorblockA{\IEEEauthorrefmark{2}National Laboratory of Wireless Communications (WiLab), CNIT, 40136 Bologna, Italy}
\IEEEauthorblockA{\IEEEauthorrefmark{3}DIEF, University of Modena and Reggio Emilia, 41125 Modena, Italy}
\IEEEauthorblockA{\IEEEauthorrefmark{4}Dipartimento di Elettronica e Telecomunicazioni, Politecnico di Torino, 10129 Torino, Italy}
\IEEEauthorblockA{\IEEEauthorrefmark{5}Dipartimento di Automatica e Informatica, Politecnico di Torino, 10129 Torino, Italy}
}


\markboth{Journal of \LaTeX\ Class Files,~Vol.~14, No.~8, August~2015}%
{Shell \MakeLowercase{\textit{et al.}}: Bare Demo of IEEEtran.cls for IEEE Transactions on Magnetics Journals}
%





\IEEEtitleabstractindextext{%
\begin{abstract} 
The automotive sector is following a revolutionary path from vehicles controlled by humans to vehicles that will be fully automated, fully connected, and ultimately fully cooperative. Along this road, new cooperative algorithms and protocols will be designed and field tested, which represents a great challenge in terms of costs. In this context, in particular, moving from simulations to practical experiments requires huge investments that are not always affordable and may become a barrier in some cases. To solve this issue and provide the community with an intermediate step, we here propose the use of 1:10 scaled cooperative, autonomous, and connected mini-cars. The mini-car is equipped with a Jetson Orin board running the open \ac{ROS2}, sensors for autonomous operations, and a Raspberry Pi board for connectivity mounting the open source \ac{OScar}. A key aspect of the proposal is the use of OScar, which implements a full ETSI \ac{C-ITS} compliant stack. The feasibility and potential of the proposed platform is here demonstrated through the implementation of a case study where the Day-1 \ac{ICW} application is implemented and validated. 
\end{abstract}
}

\maketitle

\IEEEdisplaynontitleabstractindextext

%
\IEEEpeerreviewmaketitle

\acresetall


\section{Introduction}


Robotaxis are on the road in some cities, more and more automation features are being mandated in new vehicles, and, regarding connectivity, direct \ac{V2X} connectivity is being added to the already consolidated availability of cellular communications.\footnote{In Europe, the number of vehicles equipped with ITS-G5 exceeded 2~million in 2025, while in the US the FCC has finalized at the end of 2024 new rules for the use of the ITS band by C-V2X.} Yet, most of the vehicles are still unable to share information, and the next cornerstone will be achieved once vehicles can exchange their local views and cooperate to execute joint maneuvers.

Along this path, new cooperative algorithms and protocols need to be designed and validated. Although the use of simulators is an important first step, it needs to be followed by field tests that require huge investments and may not always be affordable. Hence, an intermediate step between simulations and field tests is required. To this aim, we propose an original platform built on a 1:10 scaled car, equipped with automated functions and standard compliant communication protocols.

The idea of a scaled platform is not new, and rather 
a large number of proposals have been made recently 
as surveyed for example in \cite{10817784,10927876}. More specifically, in \cite{10817784}, more than twenty different proposals are considered and compared, which all have strengths and weaknesses, but all miss the implementation of a full standard-compliant communication protocol stack. Among them, for instance, a very low-cost solution is proposed in \cite{8794445}, consisting of 1:24 scaled mini-cars, each controlled by a Raspberry Pi Zero and connected via Wi-Fi; the limited size and cost do not allow hosting sensors like LiDARs or to install sophisticated software for control and communication. 
Additionally, a well-known platform is the slightly more expensive Duckietown proposed in \cite{7989179}, which uses a Raspberry~Pi~2 platform with a camera and Wi-Fi. As in the previous case, the limited cost and small size restrict the sensors and software that can be deployed. Among all, the setup proposed in \cite{10707232}, which was designed for the study of edge computing solutions, is the one closest to ours. The authors use indeed the same 1:10 scaled mini-cars equipped with real sensors (a LiDAR and a camera). In contrast to our proposal, they do not specifically address communication protocols and exploit standard Wi-Fi.

Hereafter, an original platform is proposed, which is based on 1:10 scaled autonomous mini-cars, equipped with an ETSI \ac{C-ITS} compliant \ac{OBU} realized using a Raspberry Pi 5 Model B device \cite{11174862}.  Remarkably, the proposed platform is, to the best of our knowledge, the first with a full standard-compliant protocol stack. The cost is around 3000~\texteuro, which is higher compared to other proposed solutions, but still sufficiently limited to make the testbed scalable.

To demonstrate the possible use of the platform, we have implemented a case study in which the Day-1 \ac{ICW} application is developed and tested using the proposed platform. 



\begin{figure}[t]
\centering
\includegraphics[width=1\columnwidth]{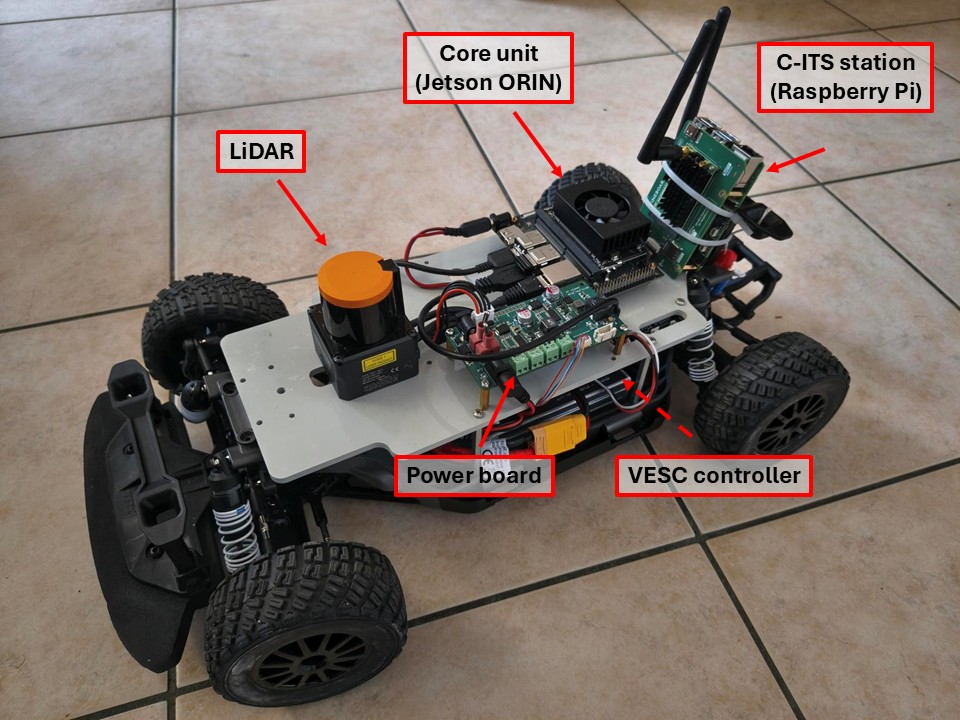}
\caption{Cooperative, connected and automated  mini-car, equipped with the low-cost OBU.} 
\label{fig:f1tenth}
\end{figure}

\section{The cooperative, connected and automated mini-car}\label{sec:plaftform}

\begin{figure}[t]
\centering
\includegraphics[width=1\columnwidth]{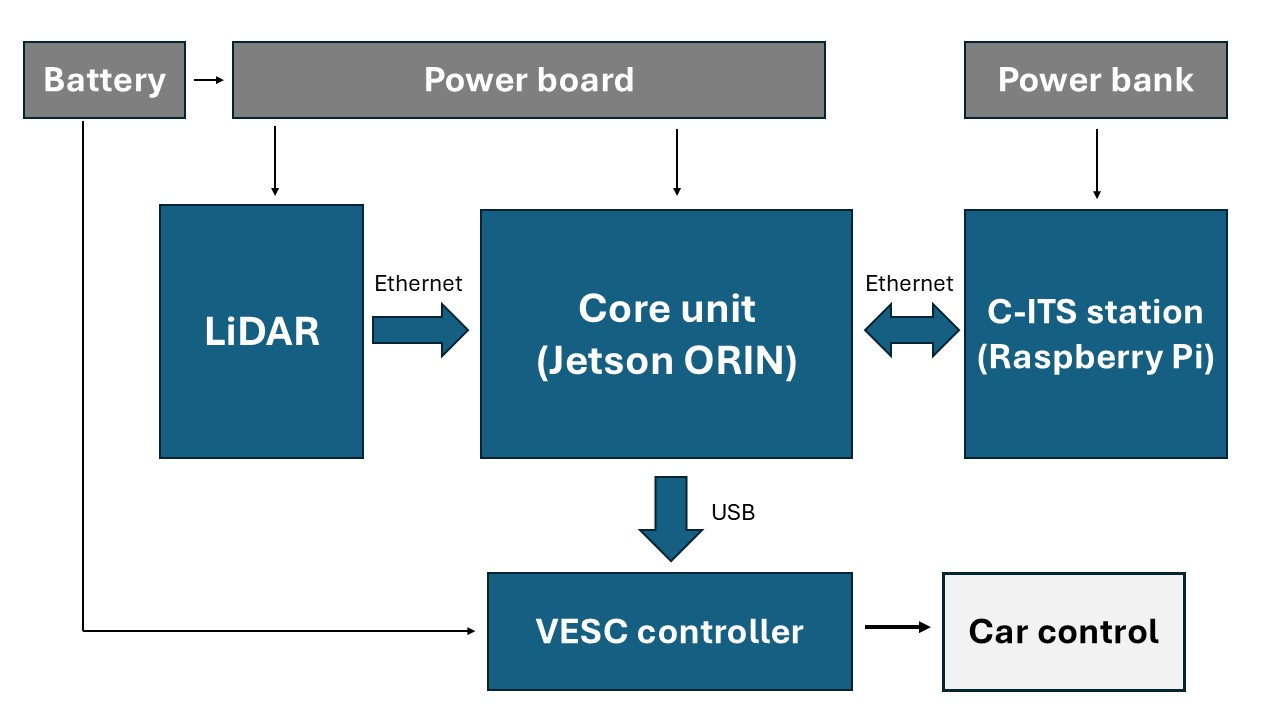}
\caption{Block scheme of the  hardware components.} \label{fig:platform}
\end{figure}

In this section, some details about the platform are provided. A picture of the platform with the main components is shown in Fig.~\ref{fig:f1tenth}, a block scheme of the main parts and their connections is illustrated in Fig.~\ref{fig:platform}, and a list of the hardware components with approximate costs is provided in Table~\ref{tab:hwlist}.

\subsection{The F1tenth mini-cars}\label{subsec:minicar}


The mini-cars are based on the Roboracer platform \cite{okelly2019f110opensourceautonomouscyberphysical}, previously known as F1tenth. This platform was created to encourage research in the field of autonomous driving. Additionally, the web site provides a basic open-source software stack and a complete guide on how to build the mini-car. Over the years, it gathered an active community that contributed to the advancement, improvement and extension of the platform. The Roboracer platform is now a mature and modular solution that offers multiple alternatives for the software components to adapt to different use-cases beyond racing.

The Roboracer autonomous platform is based on a commercially available 1/10 scaled radio-controlled 
car chassis. To enable the autonomous navigation of the vehicle, a deck was added to accommodate all the required hardware. This includes the core computing unit, sensors, power distribution and vehicle control boards. The vehicle is 
provided with a brush-less DC motor to control the speed and a servomotor for steering. These actuators are controlled with a \ac{VESC} that abstracts the motor control, exposing a serial link to the core computing unit and providing feedback data. The VESC is also equipped with a BMI160 inertial sensor, whose measurements are available via the serial interface. Finally, the main sensor for the vehicle is a 2D LiDAR rangefinder, namely a Hokuyo UST-10LX. The core computing unit that runs the autonomous navigation tasks is an NVIDIA Jetson Orin, either NX or Nano Super, which is directly connected to all sensors and actuators. The NX and Nano Super host, respectively, an 8-core or a 6-core Arm Cortex-A78AE CPU, and an embedded GPU, which can be used to speed-up computationally-intensive tasks.


The software stack is built over \ac{ROS2} \cite{ros2}, the de-facto standard framework for robotics applications. The framework provides two main abstractions: nodes, which represent software components, and topics, for inter-process communication, following a publish-subscribe paradigm. The main advantage of this approach is that the resulting software stack becomes modular, enabling the integration of ready-made software components to improve, adapt or extend the software stack.

The software components that make up the embedded software stack are the localization, the control, and the drivers for the sensors and the VESC. More specifically, the localization component is based on a particle filter~\cite{Fox2001}, a widely used algorithm in robotics state estimation based on the Monte Carlo method. The control component is instead based on the pure pursuit algorithm \cite{Wallace1985FirstRI}, a geometric control algorithm used to follow a predetermined reference trajectory.

As detailed in Table~\ref{tab:hwlist}, the cost of the hardware components to realize a mini-car, before adding the \ac{C-ITS} \ac{OBU}, is on the order of 2500-3000~\texteuro, which is mainly due to the LiDAR (more than half of the budget with current prices), the chassis, and the Jetson ORIN. Regarding the LiDAR, we use a model that allows us to comply with the RoboRacer rules; however, cheaper alternatives are in principle possible. Concerning instead the Jetson Orin, both the NX or the Nano Super can be used, with the former allowing for improved performance but at a higher cost.

\subsection{The C-ITS OBU hardware}\label{subsec:RPi}




To integrate \ac{C-ITS} \ac{V2X} capabilities into the RoboRacer autonomous platform, a compact but sufficiently powerful hardware unit and an open-source protocol stack need to be deployed. Concerning the hardware, the goal is the design of a compact low-cost solution, supporting \textit{(i)} the IEEE 802.11p standard \cite{grazia2018performance}, \textit{(ii)} the exchange of ETSI ITS-G5 messages, \textit{(iii)} dual operation as both \ac{RSU} and \ac{OBU}, \textit{(iv)} the usage of off-the-shelf components, and \textit{(v)} the suitability for rapid deployment in experimental scenarios.

Our proposed solution, also detailed in \cite{11174862}, is based on a Raspberry Pi 5 Model B single-board computer, equipped with an Arm Cortex-A76 processor, 8 GB of LPDDR4 memory, and 128 GB of flash storage. The board is powered through a 5 V / 5 A USB-C interface and provides a PCIe 2.0 connection, enabling high-speed expansion through standardized Hardware Attached on Top (HAT) modules. This design choice enables the integration of additional peripherals, such as storage devices and wireless modules, while preserving a compact and portable form factor.

For vehicular communication, a PCIe HAT is used to host a MikroTik R11e-5HnD wireless card, based on the Qualcomm Atheros AR9580 chipset and equipped with a passive heat sink. Initially intended for IEEE 802.11a/n operation, the AR9580 chipset natively supports the 5.9 GHz frequency band used for ITS communications. Access to these frequencies is enabled at the software level through the Linux \texttt{ath9k} driver, which can be extended to support IEEE 802.11p operation.


Support for ITS-G5 communications in the 5.9 GHz band is activated through a set of modular and automated configuration scripts executed at runtime. These scripts apply custom patches to the \texttt{ath9k} driver to unlock IEEE 802.11p channels and update the Linux regulatory database (\texttt{regdb}) to include the appropriate frequency allocations \cite{iandolo2025odu}.

A detailed list of the hardware components used in the proposed platform, together with their approximate costs, is reported in Table~\ref{tab:hwlist}. The overall cost of the setup is in the order of 150~\texteuro. 
This is significantly lower than the price of comparable commercial or proprietary V2X solutions, while still providing the flexibility and openness required for research, prototyping, and experimental validation.


\subsection{OScar: An ETSI C-ITS compliant protocol stack implementation}\label{subsec:Oscar}
\begin{figure}[h]
\centering
\includegraphics[width=1\columnwidth]{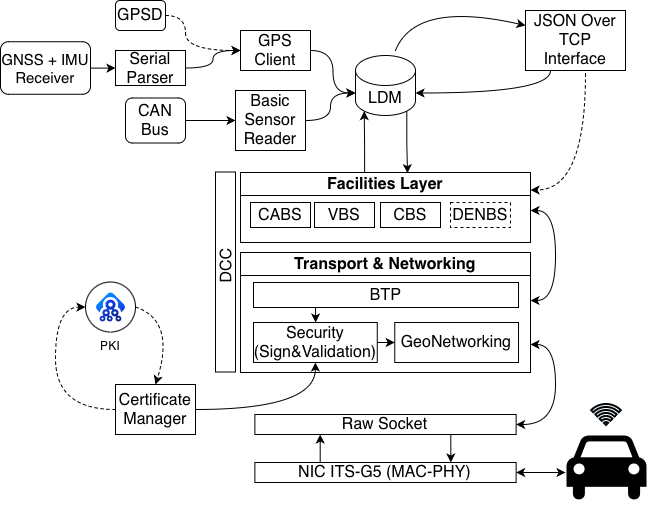}
\caption{Block diagram of the OScar stack and its main functional components}
\label{fig:OScar}
\end{figure}
The software deployed on the mini-cars is based on a fully open-source implementation of the ETSI C-ITS networking stack.
This implementation, named OScar~\cite{OScar_paper_2024}, is released under the GPLv2 license and is regularly updated with new features, message versions, and security functionalities, in line with ETSI specifications~\cite{etsi-sec}. 
The adoption of an open-source stack offers several advantages, including the ability to flexibly deploy the stack on embedded platforms such as Raspberry Pi boards without incurring licensing costs and being  able to customize the framework in all its components for research purposes. 

The internal architecture of OScar and its main functional components are illustrated in Fig.~\ref{fig:OScar}.
It follows a modular, Linux-based software stack design that can be easily deployed on low-power embedded systems.
The entire codebase can be compiled into a single executable of less than 3 MB, which can be configured either via command-line parameters or through a dedicated API that uses JSON-based data structures over TCP.
This API enables, for example, the activation of \ac{CAM} dissemination or the configuration of ETSI \ac{DCC} mechanisms.
Additional ETSI functionalities, such as DCC and secured communications, are also supported and can be enabled for extended experimental campaigns.
Through the API, it is also possible to access the contents of the internal \ac{LDM}, which maintains a dynamic representation of the road environment. 
The LDM is populated using information received from V2X messages, as well as data obtained from on-board sensors detecting other road users.
In our setup, the LDM stores the most up-to-date information about other mini-cars based on data received through \acp{CAM} version 2 messages.
This information can be retrieved by sending a JSON request that specifies a reference center position, which allows for the automatic computation of relative distances in addition to the kinematic state of surrounding mini-cars.

Beyond implementing the full set of ETSI C-ITS basic services for the transmission and reception of standardized messages, the framework also supports multiple positioning data sources. 
These comprise \textit{(i)} the Linux \texttt{gpsd} daemon, which provides processed \ac{PVT} data from a GNSS receiver; \textit{(ii)} a virtual positioning provider that replays previously collected GNSS and \ac{IMU} traces using the JSON format adopted by the TRACEN-X recording and replaying platform~\cite{tracenx_2025}; and \textit{(iii)} a highly efficient serial-parsing module that interfaces directly with GNSS receivers via a serial connection, decoding PVT data from NMEA sentences and, when available, IMU information from u-blox chipsets through UBX messages.

In our case, we leverage for positioning the cartesian coordinates expressed in the mini-car's local reference system. These are converted into NMEA sentences by a Python script running on the Jetson Orin and then forwarded to the \ac{OBU}. On the OBU, a virtual device emulates a GNSS receiver and exposes the sentences to the \texttt{gpsd} daemon, thus leveraging the standard positioning pipeline described above.

    	\begin{table}[t]
	\caption{List of hardware components and indicative prices.}\label{tab:hwlist}
		\centering 
        \scriptsize
	\begin{tabular}{m{2.2cm}p{4.4cm}p{0.9cm}}
\hline \hline
\textbf{Device} & \textbf{Details} & \textbf{Cost} \\ \hline 
\textbf{\textit{F1tenth}}\\
Chassis & Radio-controlled Traxxas mini-car & 450~\texteuro \\ 
Core control unit & Jetson Orin Nano Super/NX & 350/800~\texteuro \\ 
SD and SSD cards & Memories for the control unit & 30~\texteuro \\ 
LiDAR & 2D Hokuyo UST-10LX & 1500~\texteuro \\ 
VESC & Car mobility controller with inertial sensor & 250~\texteuro \\ 
Battery & Battery-pack & 50~\texteuro \\ 
Power board & Custom PCB for power distribution & 100~\texteuro \\ 
Small additional & ABS sheet and fasteners for assembling & 20~\texteuro \\ 
 \hline
\textbf{\textit{C-ITS radio}}\\
 Main board & Rasberry Pi 5 Model B & 80~\texteuro \\ 
Power supply & Cable-supply or power bank & 20~\texteuro \\ 
Memory & MicroSD, at least 32~GB & 	10~\texteuro \\ 
miniPCIe module & Rasp5 HAT for miniPCIe & 15~\texteuro \\ 
Fasteners & Spacers (4) and screws (4) & 5~\texteuro \\ 
Wi-Fi mPCIe & R11E-5HND ath9k & 25~\texteuro \\ 
Antenna conn. & From MMCX to SMA & 5~\texteuro \\ 
Antenna & SMA antenna for 5.9~GHz & 5~\texteuro \\ 
\hline \hline
\end{tabular}
	\end{table}

\section{Example application}\label{sec:example}

The goal of the proposed platform is to develop and validate C-ITS applications. In this section, we show a case study in which we  implemented a Day-1 application, namely the \ac{ICW}, which exploits information obtained from \acp{CAM} to increase drivers' awareness and prevent collisions at intersections. The application was implemented using Python and runs in a target car; when the car is close to an intersection and another cooperative car is approaching the same intersection, the application generates a visual alert message to warn the driver.



\begin{figure}[h]
\centering
\includegraphics[width=1\columnwidth]{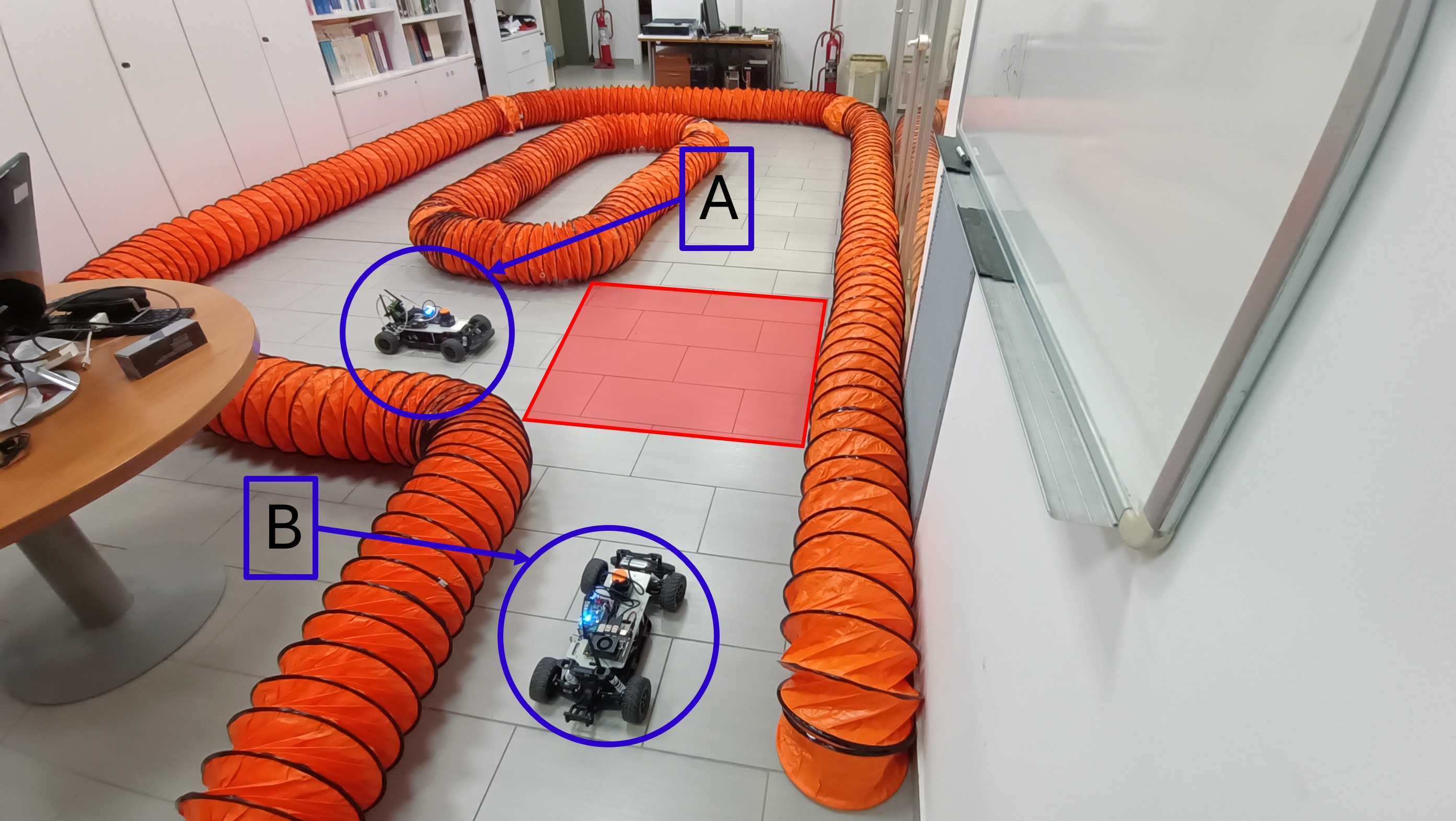}
\caption{Environment used to test the applications.}
\label{fig:track}
\end{figure}
\begin{figure}[t]
\centering
\includegraphics[width=0.8\columnwidth]{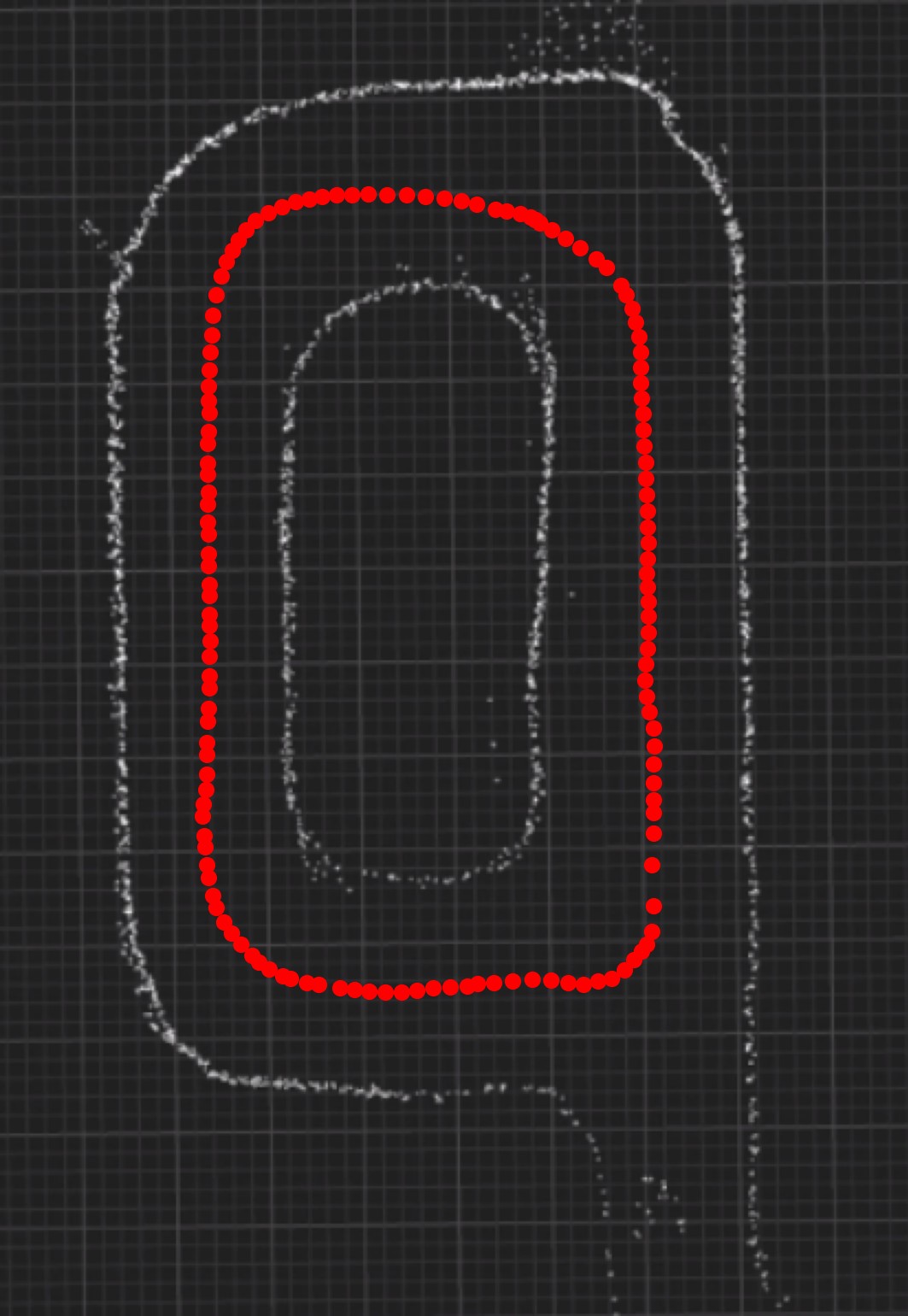}
\caption{Screenshot of the map with red points illustrating the position of mini-car A as indicated in its transmitted \acp{CAM}.}
\label{fig:map}
\end{figure}

To test the application, we set up the testing environment shown in Fig.~\ref{fig:track}. The orange flexible tubes are used to define an oval track for the mini-cars. The track is left open at one point to simulate an intersection area, highlighted in red. During the experiments, mini-car A continuously follows an oval trajectory inside the track, while mini-car B remains stationary. The autonomous guide of mini-car A is enabled by recording a map of the track using its LiDAR and by designing a reference trajectory within the mapped area. Fig.~\ref{fig:map} shows the generated map, with red points representing the coordinates indicated in the \acp{CAM} transmitted by mini-car A while following the trajectory, as received by mini-car B. Although mini-car A moves at a fixed speed of 1 m/s, the coordinate conversion mechanism described in Section~\ref{subsec:Oscar} applies a 1:10 scale (same as the mini-cars); consequently, OScar computes an effective speed higher than the real one and transmits \acp{CAM} at a frequency exceeding 1 Hz.

The ICW application is executed in a Python script on a PC connected to the \ac{OBU} of mini-car B through a socket. The script continuously sends requests to OScar to retrieve the contents of its \ac{LDM}. The received information is then processed to assess whether another vehicle (i.e., mini-car~A) is approaching the intersection area. In such a case, a warning is generated and displayed on the screen. In a real-world implementation, this warning would be shown on an in-vehicle screen or used to produce a recognizable sound, to warn the human driver of the risk of a potential collision.

\begin{figure*}[t]
\subfloat[]{\includegraphics[width=0.65\columnwidth]{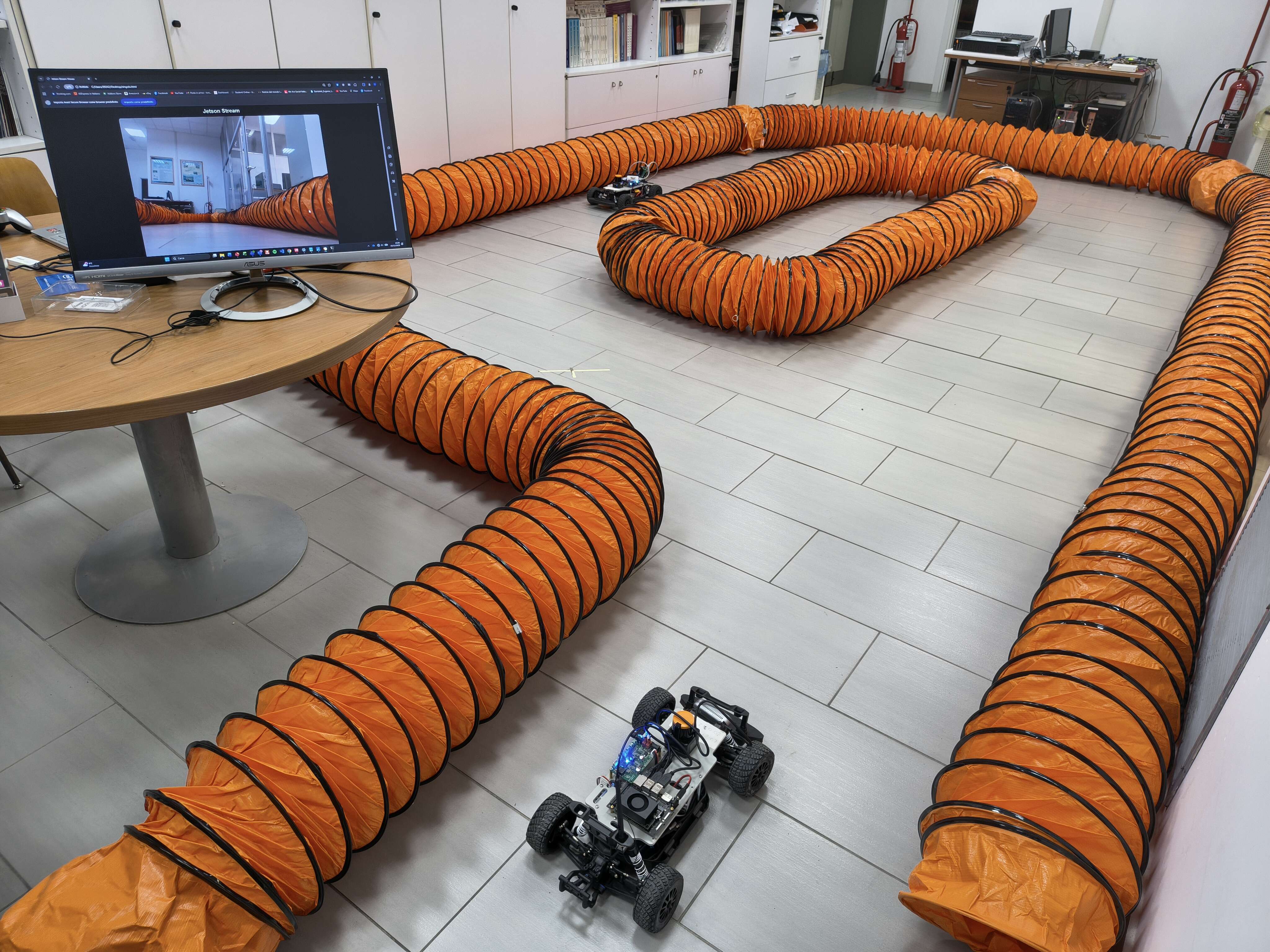}
\label{fig:esterno1}}
\hfill
\subfloat[]{\includegraphics[width=0.65\columnwidth]{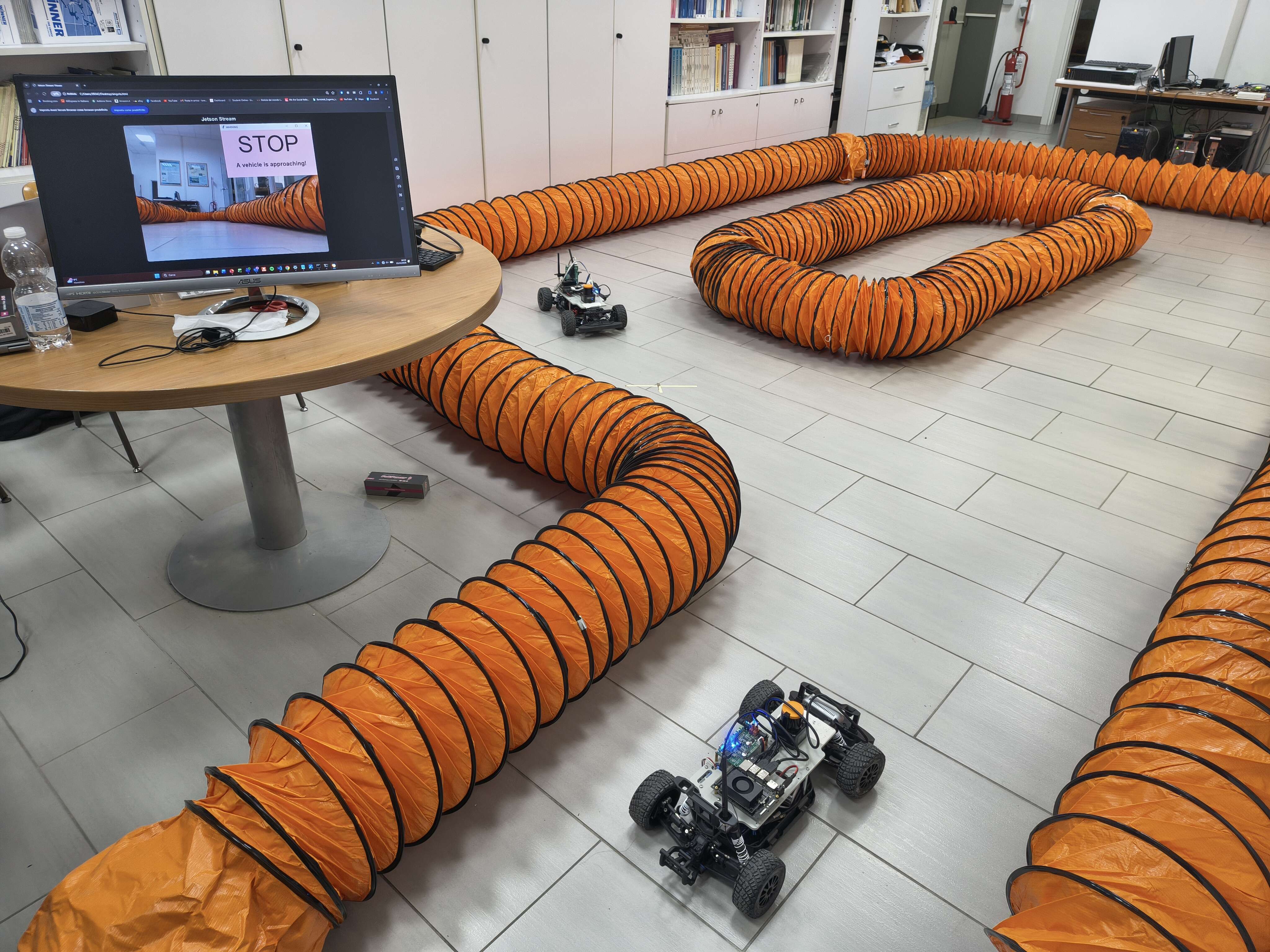}
\label{fig:esterno2}}
\hfill
\subfloat[]{\includegraphics[width=0.65\columnwidth]{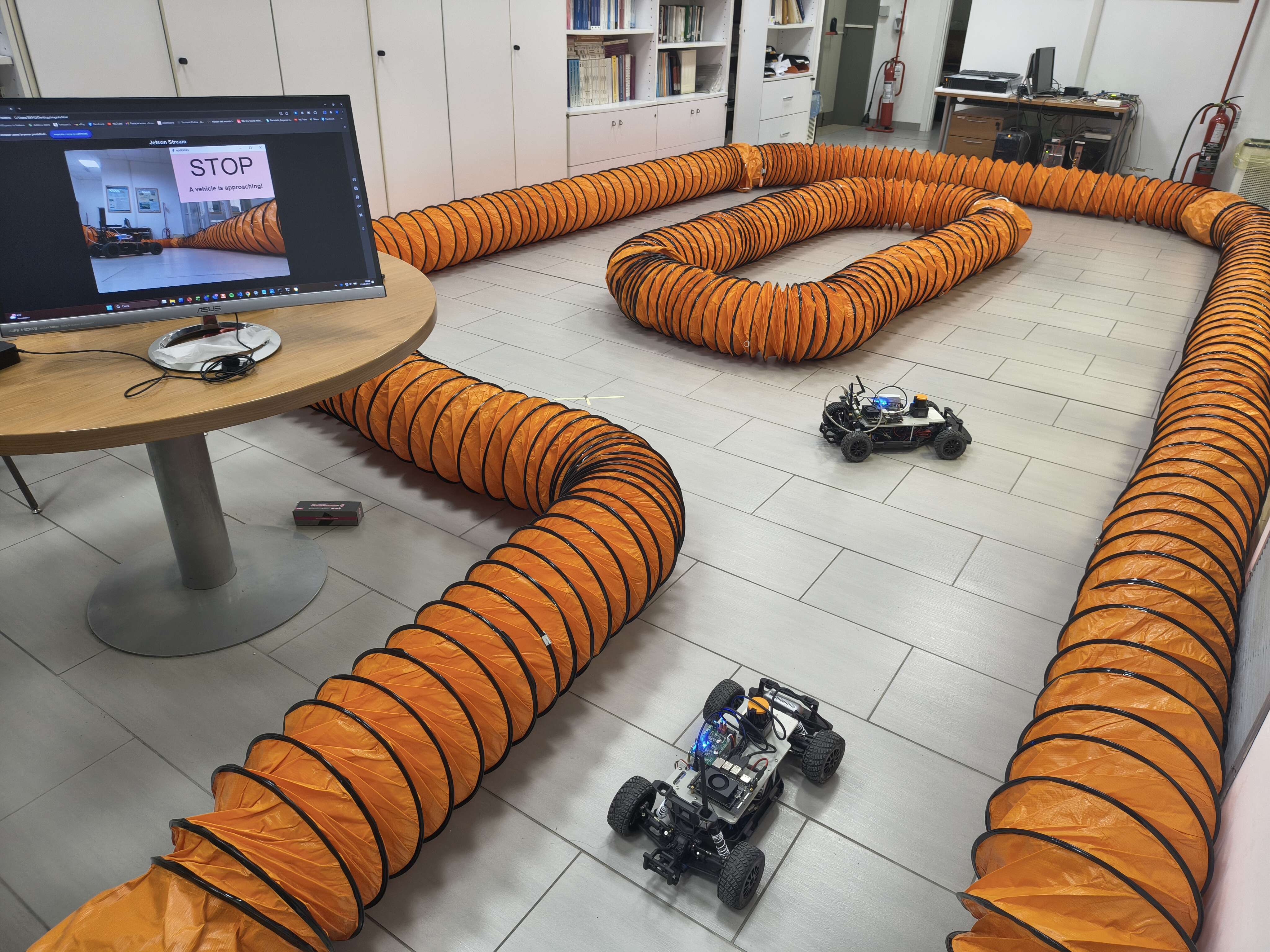}
\label{fig:esterno3}}

\subfloat[]{\includegraphics[width=0.65\columnwidth]{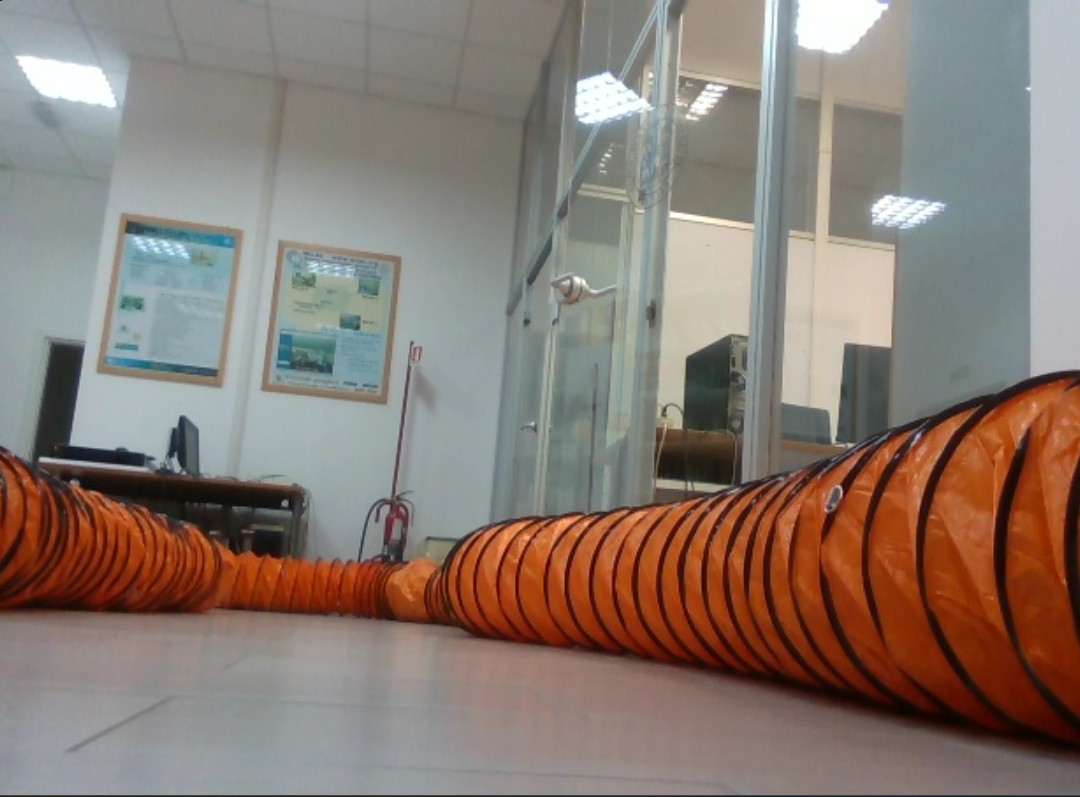}
\label{fig:pov1}}
\hfill
\subfloat[]{\includegraphics[width=0.65\columnwidth]{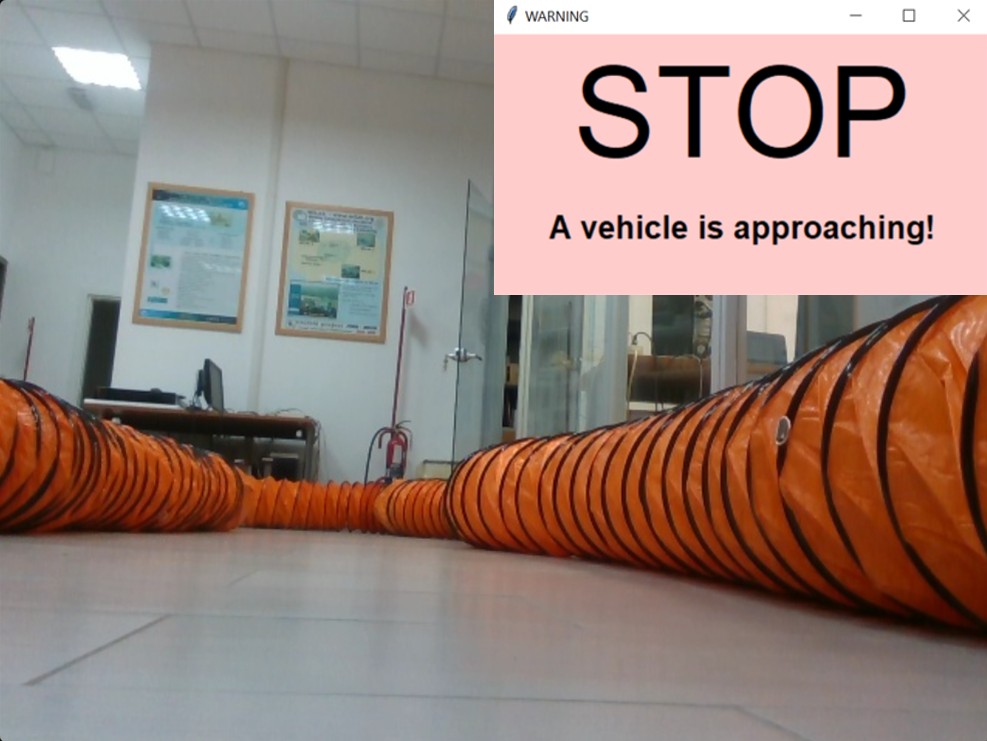}
\label{fig:pov2}}
\hfill
\subfloat[]{\includegraphics[width=0.65\columnwidth]{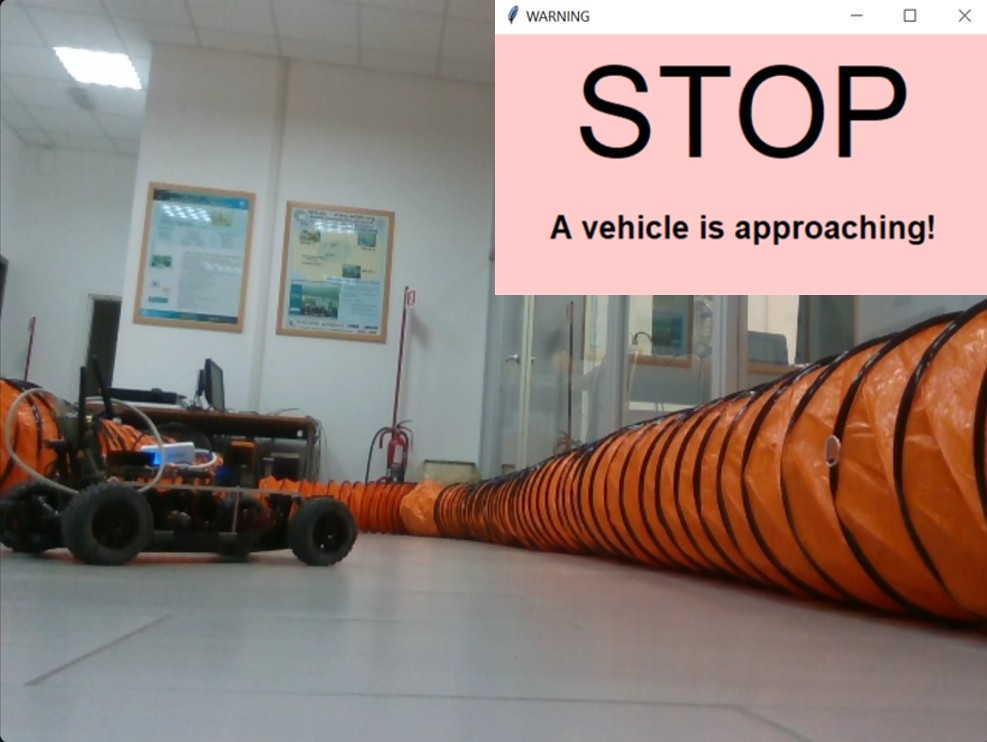}
\label{fig:pov3}}
\caption{External and on-board views of three experimental phases.
(a–-c) External camera snapshots of the three phases: mini-car A away from the intersection area, approaching the intersection area and crossing the intersection area, respectively.
(d–-f) Corresponding on-board (vehicle perspective) views for the same phases.}

\label{fig:experiment}
\end{figure*}

The experimental test used to validate the application is illustrated in Fig.~\ref{fig:experiment}, showing both external and on-board views. The on-board view is provided by a camera mounted on mini-car B, which transmits the acquired images to the PC.
Fig.~\ref{fig:esterno1} describes the setup at the start of the test: mini-car A is still away from the intersection and not visible from mini-car B, as seen in Fig.~\ref{fig:pov1}; thus, a warning for mini-car B is not yet triggered. In Fig.~\ref{fig:esterno2} instead, mini-car A approaches the intersection; Fig.~\ref{fig:pov2} shows that mini-car B is still not visually aware of the presence of mini-car A, but the warning is triggered to inform the driver of the risk of collision. Then, mini-car A actually crosses the intersection in Figs.~\ref{fig:esterno3} and~\ref{fig:pov3}. 

\section{Conclusion}\label{sec:conclusion}

In this work, a new platform for the field trial of ETSI C-ITS applications is proposed, where a 1:10 mini-car is equipped with sensors and a standard compliant \ac{OBU} using hardware with limited cost and open-source software. The applicability of the platform is demonstrated with a case study implementing a Day-1 \ac{ICW} application. The plan for future work is to apply the platform to Day-2 collective perception applications and Day-3+ maneuver coordination applications, with particular reference to intersections.


\section*{Acknowledgment} This work was partially supported by the European Union under the Italian National Recovery and Resilience Plan (NRRP) of NextGenerationEU, partnership on ``Telecommunications of the Future'' (PE00000001 - RESTART), project MoVeOver, and national research centre on mobility (CN00000023 - MOST).



\bibliographystyle{IEEEtran}  
\bibliography{biblio}

\end{document}